\documentclass[conference]{IEEEtran}
\IEEEoverridecommandlockouts
\usepackage{orcidlink}
\usepackage{cite}
\usepackage{amsmath,amssymb,amsfonts}
\usepackage{algorithmic}
\usepackage{subfig}
\usepackage{graphicx}
\usepackage{xcolor}
\usepackage[compatibility=false]{caption}
\usepackage{textcomp}
\usepackage{xcolor}
\def\BibTeX{{\rm B\kern-.05em{\sc i\kern-.025em b}\kern-.08em
    T\kern-.1667em\lower.7ex\hbox{E}\kern-.125emX}}
    
\begin{document}

\title{Evaluation of Convolutional and
Transformer-Based Detectors for Weed Detection in Tomato Plantations}
    
    

\author{
\IEEEauthorblockN{Alcides Toledo-Espinosa}
\IEEEauthorblockA{
\textit{Instituto Politécnico Nacional} \\
\textit{CIDETEC-IPN} \\
Mexico City, Mexico \\
atoledoe2600@alumno.ipn.mx
}
\and
\IEEEauthorblockN{Gerardo Antonio Alvarez-Hernandez}
\IEEEauthorblockA{
\textit{Instituto Politécnico Nacional} \\
\textit{CIDETEC-IPN} \\
Mexico City, Mexico \\
galvarezh1400@alumno.ipn.mx
}
\and
\IEEEauthorblockN{Ángel Eduardo Zamora-Suárez}
\IEEEauthorblockA{
\textit{Instituto Politécnico Nacional} \\
\textit{UPIBI-IPN} \\
Mexico City, Mexico \\
azamora@ipn.com
}
\and
\IEEEauthorblockN{Miguel Hernandez-Bolaños}
\IEEEauthorblockA{
\textit{Instituto Politécnico Nacional} \\
\textit{CIDETEC-IPN} \\
Mexico City, Mexico \\
mbolanos@ipn.mx
}
\and
\IEEEauthorblockN{Juan Irving Vasquez}
\IEEEauthorblockA{
\textit{Instituto Politécnico Nacional} \\
\textit{CIDETEC-IPN} \\
Mexico City, Mexico \\
jvasquezg@ipn.mx
}
}

\maketitle

\begin{abstract}
This paper presents a comparative evaluation of convolutional and transformer-based object detection architectures for early weed detection in tomato plantations. Representative models from each paradigm are considered, including YOLOv26-nano, a recent variant of the YOLO family, and RT-DETR Large and RF-DETR Medium as transformer-based architectures. The evaluation was conducted on the GROUNDBASED\_WEED dataset, considering six weed classes and an additional category corresponding to unidentified plants, which allowed for the assessment of performance in terms of detection accuracy and computational efficiency using metrics such as precision, recall, average precision, and inference speed, as well as non-parametric statistical tests.

The results highlight a clear trade-off between efficiency and contextual modeling: CNN-based detectors achieve high performance at a lower computational cost, while transformer-based approaches offer better global context capture at the expense of higher resource demands. These results provide practical criteria for model selection in precision agriculture applications.
\end{abstract}


\section{Introduction}

Agricultural systems are undergoing a profound transformation driven by the need to simultaneously increase productivity and ensure environmental sustainability. Precision agriculture addresses this challenge through site-specific management strategies that exploit the spatial and temporal variability of fields, integrating advanced technologies such as robotics, computer vision, and artificial intelligence \cite{kamilaris2018deep, liakos2018machine}. Among the problems that precision agriculture must address, weed management stands out as particularly critical: weeds compete with crops for nutrients, water, and sunlight, cause significant yield losses, and serve as hosts for pests and diseases \cite{heap2014global, chauhan2024weed, dentika2021weeds}. Traditional control strategies — uniform herbicide application and mechanical removal — are inefficient, environmentally unsustainable, and drive the emergence of herbicide-resistant species \cite{heap2014global, kazinczi2023herbicide}, underscoring the need for selective interventions such as site-specific spraying \cite{slaughter2008autonomous}. Enabling such precision, however, requires robust automated systems capable of reliably distinguishing crops from weeds under complex field conditions.
In response to this need, deep learning has substantially advanced computer vision for agricultural applications. Convolutional neural networks (CNNs) have demonstrated strong performance in classification, detection, and segmentation tasks, consistently outperforming handcrafted-feature approaches \cite{dos2017weed, milioto2018real}. However, their inherently localized receptive fields limit the capture of long-range dependencies, a critical shortcoming in fine-grained scenarios such as early-stage crop–weed discrimination \cite{qu2024deep}. Transformer-based architectures — including Vision Transformers (ViTs) \cite{dosovitskiy2020image}, DETR \cite{carion2020end}, and self-supervised models such as DINOv2 \cite{oquab2023dinov2} — have emerged as promising alternatives by capturing global contextual relationships via self-attention. Nevertheless, their high demand for large-scale datasets and computational resources limits their applicability under realistic agricultural conditions, motivating the exploration of hybrid CNN–transformer approaches.

Despite these advances, early-stage weed detection remains a particularly challenging problem. During early growth stages, crops and weeds exhibit high morphological similarity, high intra-class variability, and low inter-class separability. Compounding this difficulty, agricultural environments introduce significant variability in illumination, soil texture, plant density, and occlusion, which limits the robustness and generalization capability of existing models \cite{qu2024deep}. Critically, most studies have been conducted under controlled conditions, limiting adaptability across diverse field scenarios; furthermore, well-annotated, diverse datasets remain scarce, and systematic comparisons between convolutional and transformer-based architectures within unified experimental frameworks are still lacking. 

Motivated by these gaps, this work extends the study presented by Adrià et al. \cite{gomez2025spatio} by considering a more detailed classification scheme with seven classes instead of the five categories previously evaluated. In addition, more recent detection architectures are incorporated, including YOLOv26-nano and RF-DETR Medium, complementing earlier evaluations based on YOLOv8--YOLOv10 and RT-DETR. Under this framework, CNN-based and transformer-based approaches are systematically benchmarked on the GROUNDBASED\_WEED dataset under realistic field conditions in tomato plantations. 


\section{Materials and Methods}\label{Methods}

This section presents the methodological framework for evaluating convolutional and transformer-based object detection architectures for early-stage weed detection under real-field conditions. Emphasis is placed on realistic datasets, standardized protocols, and reproducible implementation.

The methodology is organized into five components: (\ref{dataset_seccion}) dataset description, (\ref{Convolution_seccion}) convolution-based detectors, (\ref{transformes_seccion}) transformer-based detectors, (\ref{design_seccion}) experimental design, and (\ref{implentation_seccion}) implementation. All models were trained and evaluated using the same dataset partitions, preprocessing pipeline, evaluation metrics, and hardware setup. To minimize sources of variability unrelated to architectural differences, training hyperparameters —  including the number of epochs, image resolution, and optimization strategy — were standardized across all models,  ensuring that observed performance differences are attributable to architectural design rather than experimental conditions.

Model robustness and generalization are evaluated under realistic field conditions, including class imbalance, annotation noise, and high morphological similarity between crops and weeds.

\subsection{Dataset}\label{dataset_seccion}

The public GROUNDBASED\_WEED dataset was used for evaluation. It consists of high-resolution RGB images ($5184 \times 3888$ pixels) captured in commercial tomato fields in Santa Amalia, Spain, using a Canon PowerShot SX540 HS camera \cite{wu2019design}. The dataset reflects realistic agricultural conditions, incorporating natural variability in illumination and scene composition.

Images were acquired from a bird’s-eye perspective at a height of 1–1.5 m, yielding a ground sampling distance of approximately 0.05 cm/pixel. This high resolution supports fine-grained analysis of plant morphology, which is critical for early-stage discrimination \cite{murad2023weed}.

The dataset focuses on early growth stages, corresponding to the critical intervention window where morphological similarity between crops and weeds is highest, making classification particularly challenging \cite{slaughter2008autonomous}.

Six weed classes and an additional category corresponding to unidentified plants were annotated following the coding system defined by the European and Mediterranean Plant Protection Organization (EPPO) see Fig.\ref{fig:dataset_examples}, which assigns standardized identifiers to plant species used in agricultural and phytosanitary applications:

\begin{figure}[ht]
    \centering
    \subfloat[CYPRO]{
        \includegraphics[width=0.3\linewidth]{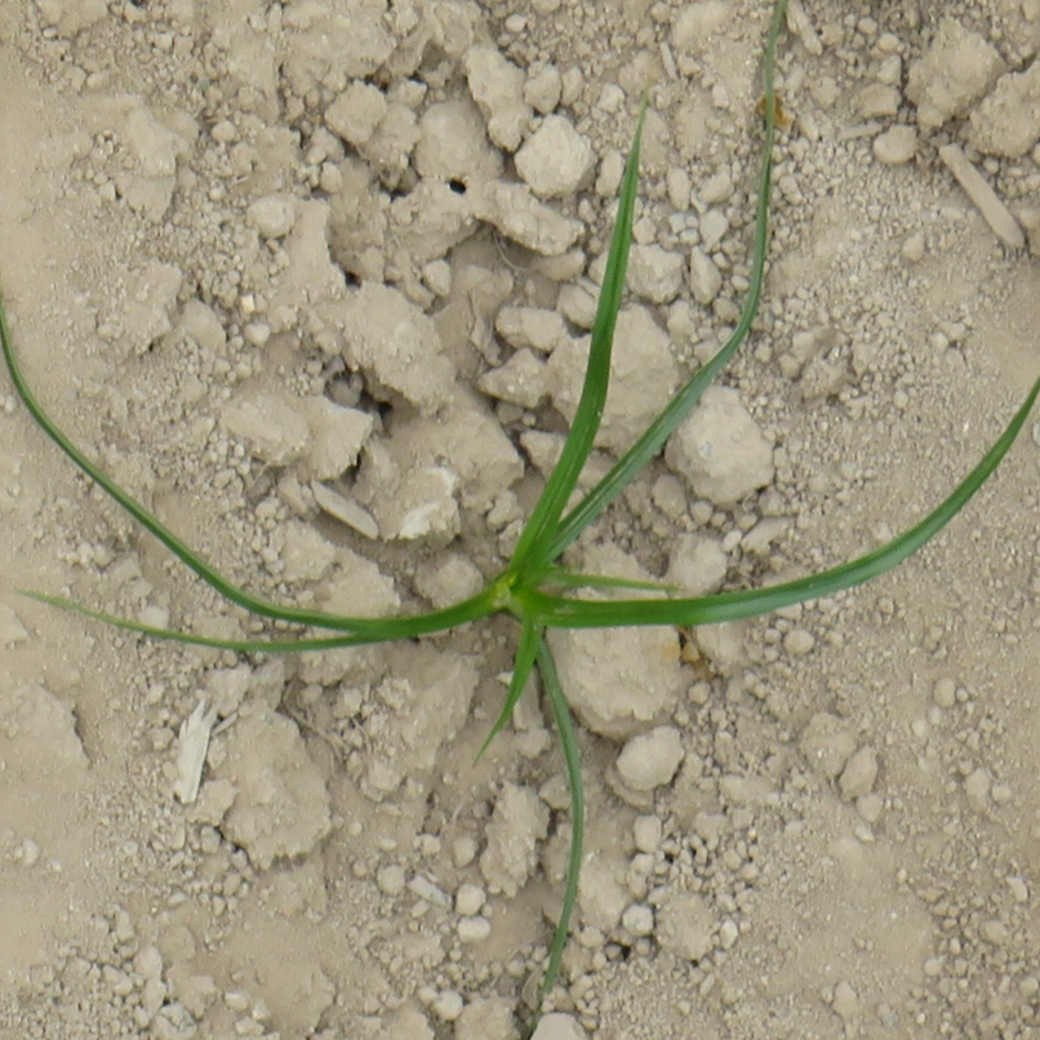}
    }
    \hfill
    \subfloat[SOLNI]{
        \includegraphics[width=0.3\linewidth]{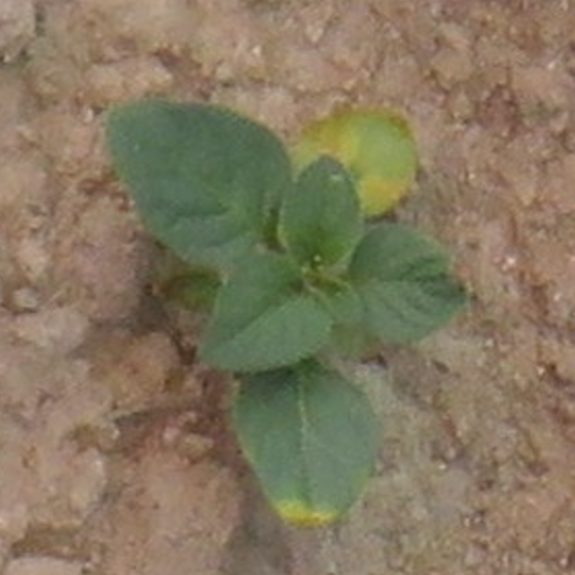}
    }
    \hfill
    \subfloat[LYPES]{
        \includegraphics[width=0.3\linewidth]{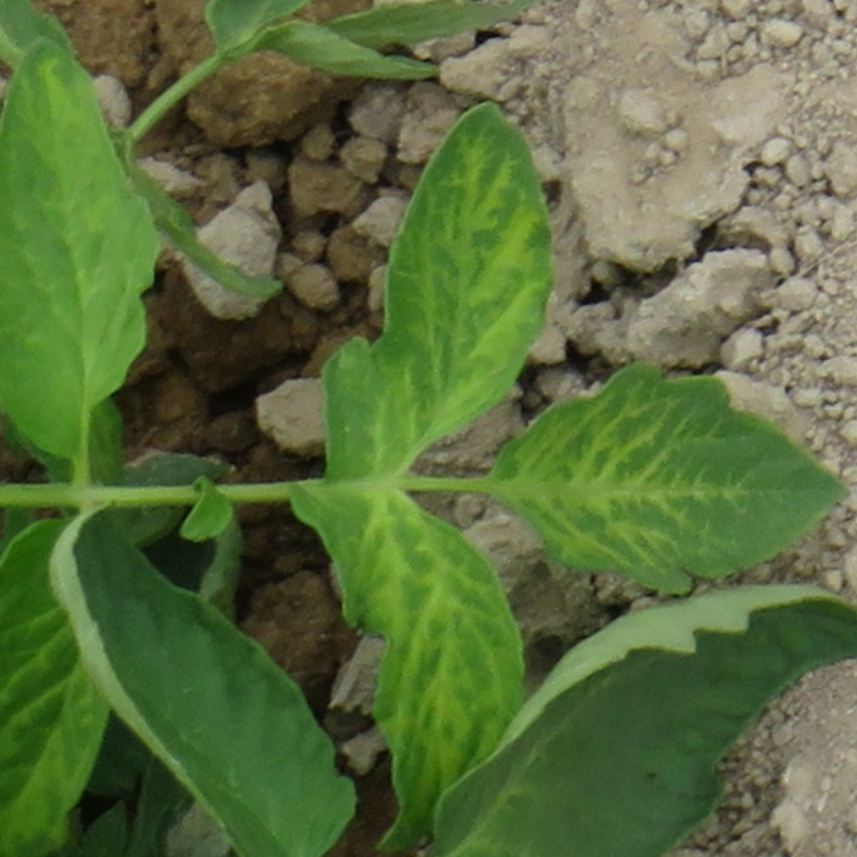}
    }
    \hfill
    \subfloat[ECHCG]{
        \includegraphics[width=0.3\linewidth]{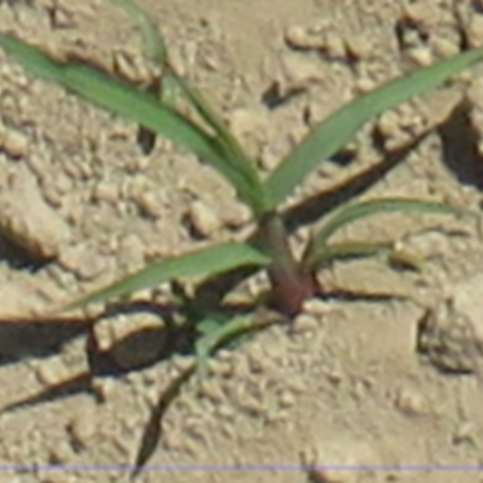}
    }
    \hfill
    \subfloat[SETVE]{
        \includegraphics[width=0.3\linewidth]{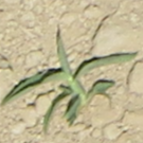}
    }
    \hfill
    \subfloat[POROL]{
        \includegraphics[width=0.3\linewidth]{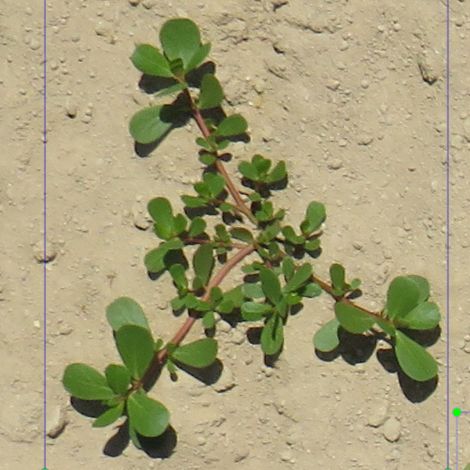}
    }

   \caption{
    Examples of dataset classes: 
    LYPES: \textit{Solanum lycopersicum} L.; 
    SOLNI: \textit{Solanum nigrum} L.; 
    ECHCG: \textit{Echinochloa crus-galli} L.; 
    CYPRO: \textit{Cyperus rotundus} L.; 
    SETVE: \textit{Setaria verticillata} L.; 
    POROL: \textit{Portulaca oleracea} L.
    }
    \label{fig:dataset_examples}
\end{figure}

\begin{figure}
    \centering
    \includegraphics[width=1\linewidth]{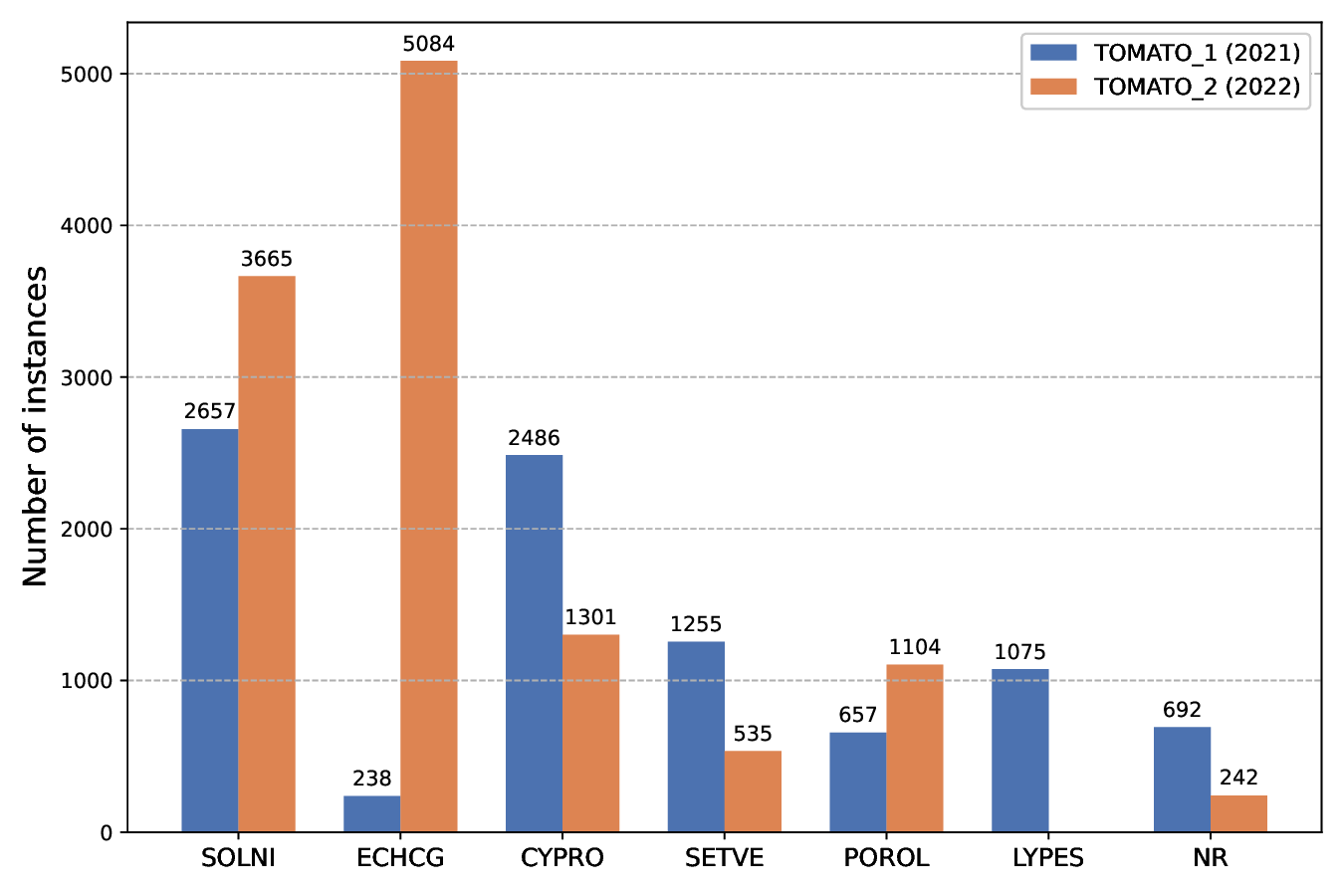}
    \caption{Distribution of annotated instances in the GROUNDBASED\_WEED dataset, highlighting class imbalance and ambiguous samples.}
    \label{fig:weed_stats}
\end{figure}

The dataset exhibits class imbalance, as summarized in Figure~\ref{fig:weed_stats}, reflecting field variability in which certain species appear more frequently than others \cite{he2009learning}. Additionally, annotation inconsistencies were identified. In particular, the LYPES class shows missing labels in the TOMATO\_2 subset despite visual presence, indicating label noise \cite{frenay2013classification}. Approximately 4.45\% of samples are labeled as NR due to ambiguity or insufficient visual information \cite{MORENO2025112249}.

These characteristics make GROUNDBASED\_WEED a representative benchmark for evaluating model robustness and generalization, bridging the gap between controlled experimental settings and practical agricultural deployment.

\subsection{Convolution-Based Detectors}\label{Convolution_seccion}

Convolutional Neural Network (CNN)-based object detectors capture a dataset's original complexity by training models to achieve strong performance and computational efficiency. These methods are typically categorized into two-stage and one-stage detectors, offering different trade-offs between accuracy and speed.

Two-stage detectors, such as the R-CNN family \cite{girshick2014rich,ren2015faster}, achieve high accuracy thresholds, but their sequential processing limits real-time applicability. In contrast, one-stage detectors directly predict bounding boxes and class probabilities in a single forward pass, enabling faster inference. Among these, the YOLO family \cite{redmon2016you} has become a dominant approach for real-time object detection.

Recent advancements, such as YOLOv26-nano \cite{sapkota2025yolo26}, further improve efficiency and detection performance by simplifying the detection pipeline and optimizing feature representation, making them suitable for real-time applications.

CNN-based detectors are particularly effective at capturing local spatial features, such as edges, textures, and shapes, which are essential for object discrimination. However, their limited ability to model global contextual relationships may reduce performance in fine-grained scenarios, such as early-stage crop–weed detection.

\subsection{Transformer-Based Detectors}\label{transformes_seccion}

Transformer-based architectures have introduced a new paradigm in object detection by leveraging self-attention mechanisms to capture global contextual relationships within images. This approach was introduced by the Detection Transformer (DETR) \cite{carion2020end}, which formulates detection as a direct set prediction problem within an end-to-end framework.

However, the original DETR suffers from slow convergence and limited performance on small objects due to insufficient multi-scale feature representation. To address these limitations, several improved variants have been proposed.

Among them, RT-DETR \cite{zhao2024detrs} introduces a hybrid architecture that combines convolutional backbones with transformer-based attention modules, enabling a better balance between accuracy and real-time performance. As a result, it has become one of the first transformer-based detectors to compete with state-of-the-art CNN-based models in both accuracy and latency. In contrast, RF-DETR \cite{rfdetr2025} adopts a fully transformer-based design, built on a pretrained DINOv2 vision transformer backbone and a lightweight DETR decoder. Rather than relying on convolutional feature extraction, RF-DETR prioritizes domain adaptability and generalization to diverse datasets, becoming the first real-time detector to surpass 60 mAP on the COCO benchmark. While RT-DETR achieves efficiency through its CNN–transformer hybrid design, RF-DETR pushes the boundaries of pure transformer-based detection, representing two distinct strategies within the same architectural family.

Transformer-based detectors are particularly effective in modeling long-range dependencies and global context, which is critical in fine-grained tasks such as early-stage crop–weed detection. However, they typically require higher computational resources and larger datasets, limiting their applicability in real-time or resource-constrained environments.

Overall, these models provide improved contextual modeling while highlighting trade-offs between accuracy and computational efficiency.

\subsection{Design of Experiment}\label{design_seccion}

This study presents a comparative evaluation of object detection architectures based on CNNs and transformers for early weed detection in agricultural field environments. The YOLOv26-nano model was selected as a representative of CNN-based detectors due to its lightweight design, which is particularly relevant for agricultural applications where deployment on embedded systems is sought. Similarly, lightweight variants of transformer-based architectures, specifically RF-DETR Medium and RT-DETR Large (based on the DETR framework), were selected to ensure a fair comparison under practical computational constraints. The objective is to evaluate detection accuracy, generalization, and computational efficiency in challenging agricultural environments.

All models were trained using a standardized protocol to ensure a fair comparison. Training was performed for up to 200 epochs with early stopping (patience = 20). The AdamW optimizer was used with a learning rate of $1 \times 10^{-5}$, and a batch size of 4 was selected to balance memory limitations and training stability. These hyperparameters were adopted following the configuration reported by Adrià et al. \cite{gomez2025spatio}, who used the same dataset and addressed a similar weed detection task.

The full complexity of the GROUNDBASED\_WEED dataset was preserved by retaining all 7 annotated classes, enabling rigorous evaluation in a fine-grained classification setting.

To reduce computational cost, a multi-scale resizing strategy was applied, evaluating input resolutions of 640, 1080, and 2160 pixels. This enables the analysis of the trade-off between detection accuracy and computational efficiency.

The dataset was partitioned into training, validation, and test sets with an 80\%–10\%–10\% split, yielding 972 images for training and 122 images each for validation and testing. This partitioning strategy ensures that model training, hyperparameter tuning, and performance evaluation are conducted on independent subsets, supporting an unbiased assessment of model performance.

Performance was evaluated using standard object detection metrics, including precision, recall, and mean average precision (mAP@50 and mAP@50:90). An intersection-over-union (IoU) threshold of 0.5 and a confidence threshold of 0.3 were used.

To provide a more objective comparison between architectures, the experimental results were analyzed using two nonparametric statistical tests, which are described in detail in Section \ref{III-A}

\subsection{Implementation}\label{implentation_seccion}

All experiments were conducted in a reproducible environment using Python 3.10 with Conda for dependency management. Models were implemented in PyTorch with GPU acceleration. The Ultralytics library was used for CNN-based detectors (YOLOv26), while transformer-based models (RT-DETR) were implemented using their official PyTorch frameworks.

Experiments were performed on a workstation with an Intel Core i7 processor, 32 GB RAM, and an NVIDIA RTX 2080 Super GPU (8 GB), with CUDA and cuDNN enabled for acceleration.

Reproducibility was ensured through fixed random seeds, identical data splits, and consistent preprocessing pipelines. Model checkpoints were selected based on validation performance, and mixed precision (FP16) was used when supported to reduce memory usage and accelerate training.

Training times ranged from 6 to 8 hours per model, with transformer-based models requiring higher computational resources, while CNN-based models showed faster convergence and lower overhead.

\begin{figure*}[t]
    \centering
    \subfloat[Ground truth]{
        \includegraphics[trim={0 700pt 0 0},clip,height=0.25\textheight]{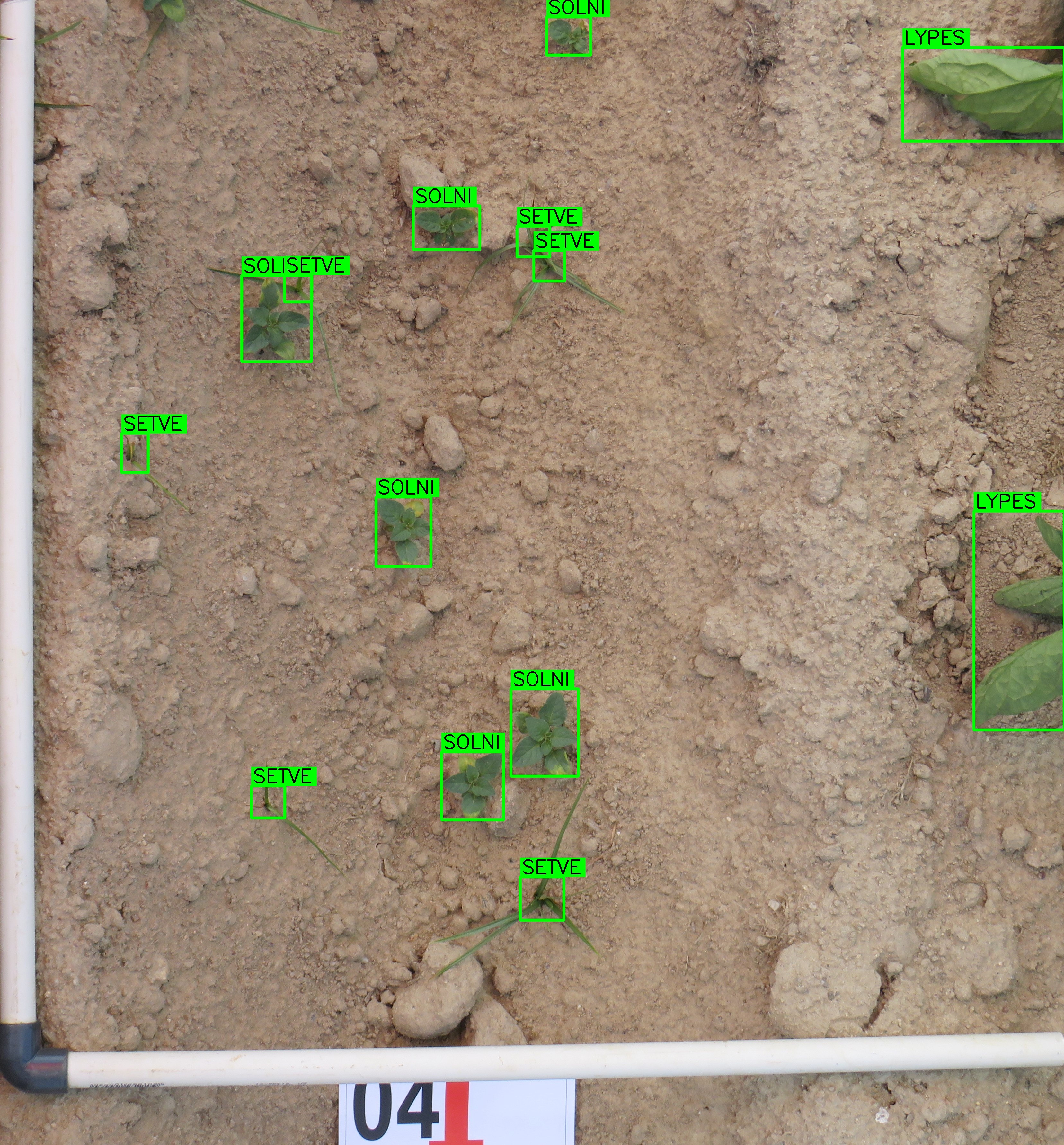}
        \label{fig:inf1}
    }
    \hspace{0.25cm}
    \subfloat[Inference YOLOv26-nano]{
        \includegraphics[trim={0 700pt 0 0},clip,height=0.25\textheight]{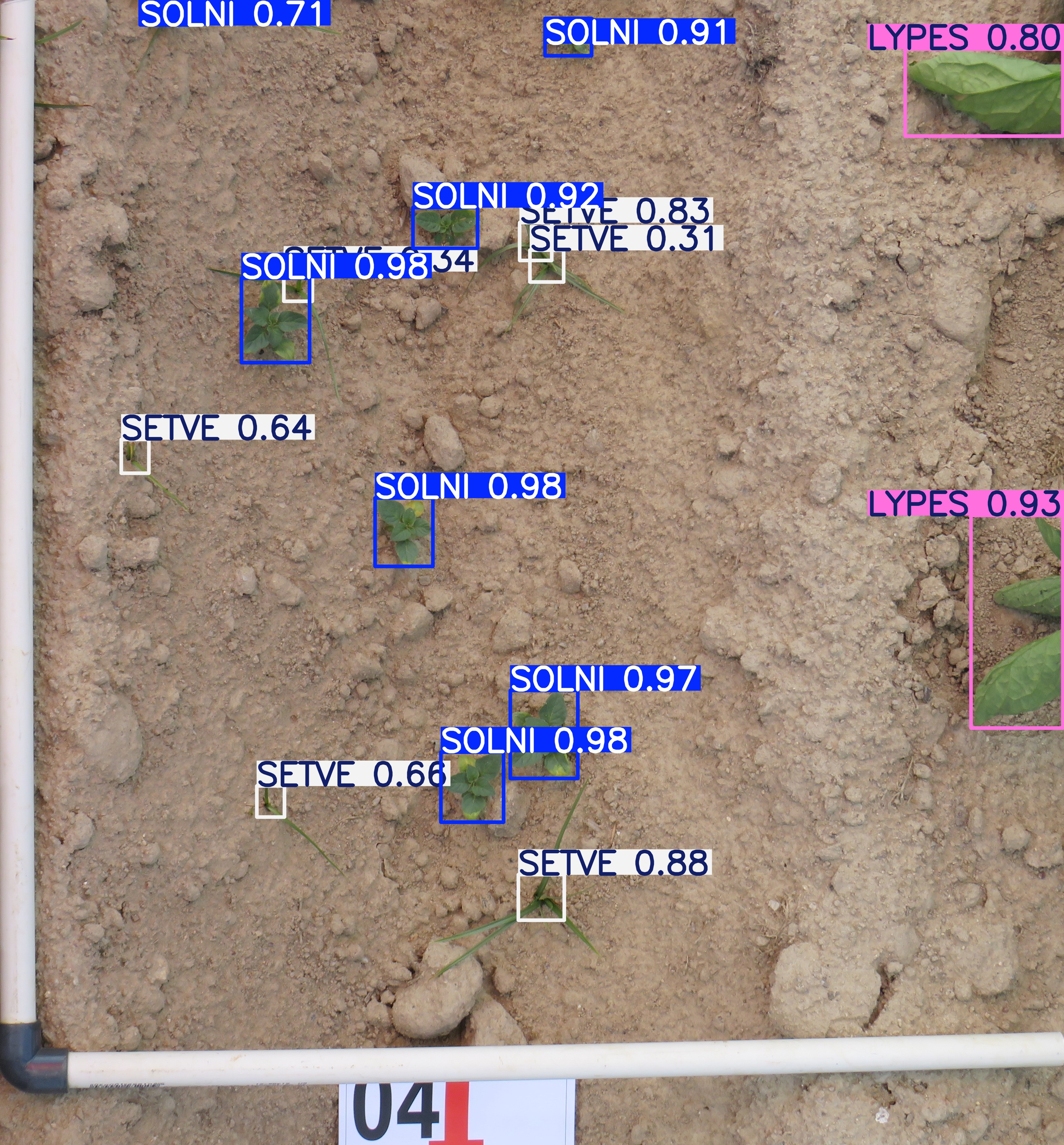}
        \label{fig:inf2}
    }
    
    \subfloat[Inference RF-DETR Medium]{
        \includegraphics[trim={0 700pt 0 0},clip,height=0.25\textheight]{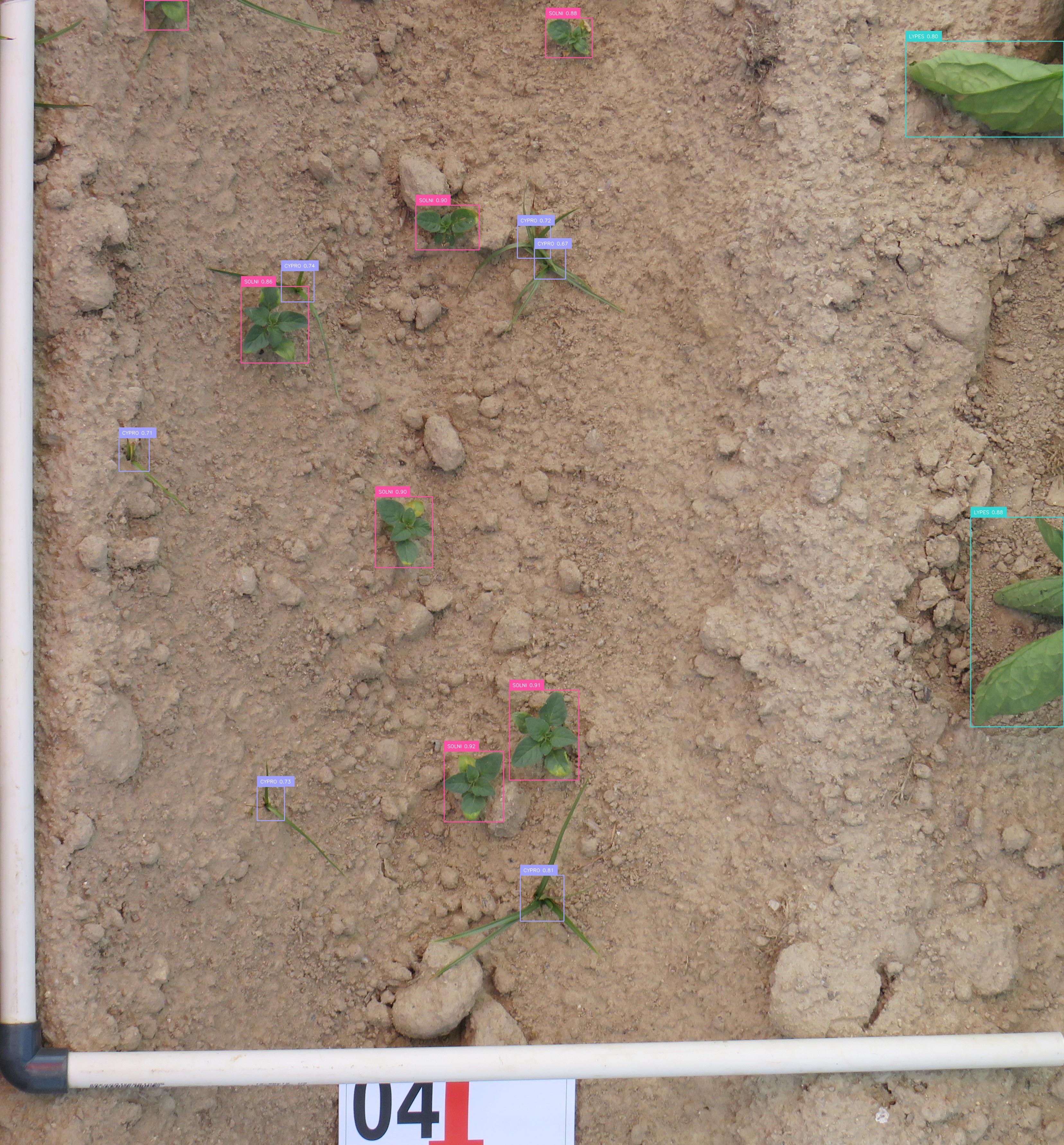}
        \label{fig:inf3}
    }
    \hspace{0.25cm}
    \subfloat[Inference RT-DETR large]{
        \includegraphics[trim={0 700pt 0 0},clip,height=0.25\textheight]{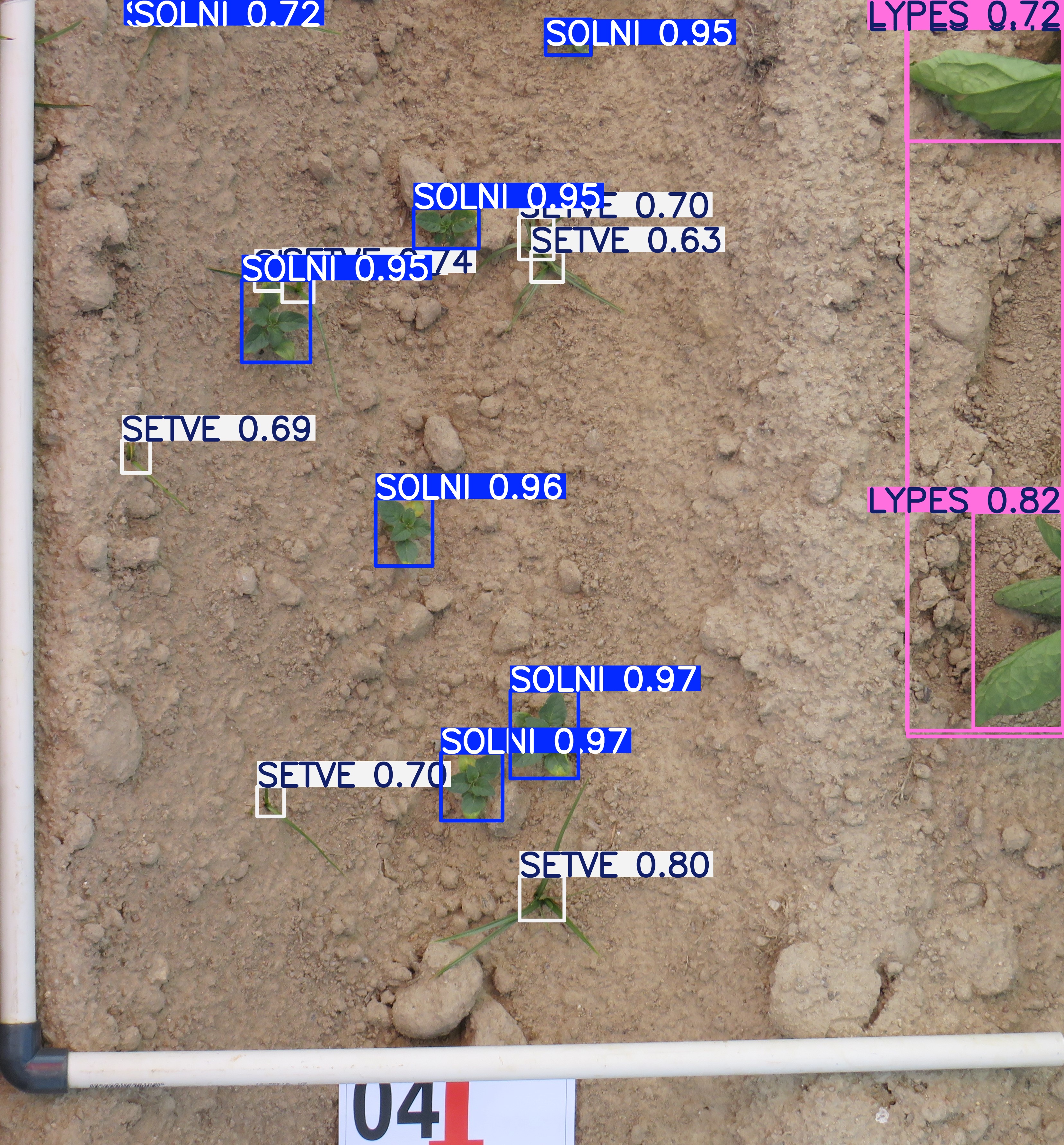}
        \label{fig:inf4}
    }
    \caption{Representative inference examples generated by YOLOv26-nano, RF-DETR Medium, RT-DETR large on the GROUNDBASED\_WEED dataset.}
    \label{fig:yolo_inference}
\end{figure*}

\section{Results and Discussion}
\label{Results}

To evaluate the performance of the selected detection frameworks, a series of experiments was conducted using the hyperparameters specified in Section \ref{design_seccion}. The evaluated models YOLOv26-nano, RT-DETR Large and RF-DETR medium were chosen as representative state-of-the-art approaches. Specifically, this selection contrasts a highly optimized convolution-based model (YOLOv26-nano) with two transformer-based architectures (RT-DETR Large and RF-DETR Medium) to analyze the trade-offs between detection accuracy and computational overhead.

As summarized in Table \ref{tab:comparison_detection_models}, YOLOv26-nano and RT-DETR Large exhibit comparable performance regarding Precision, Recall, and mAP@50:90, especially at higher resolutions. The robustness of YOLOv26-nano stems from its efficient single-stage design, which leverages hierarchical spatial feature extraction to detect multi-scale objects at a low computational cost. Conversely, RT-DETR Large integrates self-attention mechanisms to model global context, potentially enhancing object localization in complex scenes.

\begin{table*}[htp]
\centering
\caption{Performance comparison of YOLOv26-nano, RT-DETR Large, and RF-DETR Medium}
\label{tab:comparison_detection_models}
\begin{tabular}{c|c|c|c|c|c|c}
\hline
\textbf{Model} & \textbf{Experiment} & \textbf{Image Size} & \textbf{Precision} & \textbf{Recall} & \textbf{mAP@50} & \textbf{mAP@50:90} \\
\hline
YOLOv26-nano & 1 & 640  & 0.716 & 0.595 & 0.665 & 0.468 \\
YOLOv26-nano & 2 & 1080 & 0.721 & 0.648 & 0.704 & 0.513 \\
YOLOv26-nano & 3 & 2160 & 0.728 & \textbf{0.692} & 0.729 & \textbf{0.533} \\
\hline
RT-DETR Large & 1 & 640  & 0.710 & 0.585 & 0.626 & 0.416 \\
RT-DETR Large  & 2 & 1080 & 0.747 & 0.657 & 0.702 & 0.487 \\
RT-DETR Large & 3 & 2160 &\textbf{ 0.760} & 0.690 & \textbf{0.730} & 0.510 \\
\hline
RF-DETR Medium  & 1 & 640  & 0.639 & 0.527 & 0.563 & 0.374 \\
RF-DETR Medium & 2 & 1080 & 0.672 & 0.591 & 0.632 & 0.438 \\
RF-DETR Medium & 3 & 2160 & 0.684 & 0.621 & 0.657 & 0.459 \\
\hline
\end{tabular}
\end{table*}

However, although RT-DETR Large shows marginal improvements in $mAP@{50}$ at high resolutions, these gains are not statistically significant compared to YOLOv26-nano. This suggests that for this specific agricultural application, the global modeling capabilities of Transformers do not provide a substantial advantage over convolutional representations. This behavior is likely due to the inherent nature of the target objects, which are characterized by small scales and relatively uniform spatial distributions.

Furthermore, it is crucial to note that the slight performance gains in RT-DETR Large entail a considerable increase in computational complexity. Due to its Transformer-based backbone, RT-DETR Large requires higher memory allocation and yields longer inference times, which may hinder its feasibility for real-time deployment on embedded robotic platforms. In the case of {RF-DETR Medium}, the model followed a similar trend but consistently underperformed across all metrics, potentially due to the sensitivity of attention-based architectures to dataset scale and hyperparameter tuning.

It is essential to contextualize these findings with respect to prior studies, such as the work of \cite{gomez2025spatio}, which used the same dataset and reported higher performance metrics. However, it is important to note that those results were obtained under a simplified classification setting involving only five classes, whereas the present study considers a more comprehensive scheme with seven classes. By preserving the complete set of labels, this work introduces a more challenging and realistic evaluation scenario.

Lastly, increasing the input resolution from 640 to 2160 pixels yielded only negligible improvements. This suggests that the challenge lies in the dataset's intrinsic characteristics: since the target objects occupy a very small fraction of the image, higher resolutions do not necessarily yield more discriminative features.

\subsection{Statistical Analysis} \label{III-A}

To assess whether observed performance differences among architectures are statistically significant, non-parametric hypothesis tests were employed. Non-parametric methods were selected because the number of experimental observations per model is limited ($n = 3$), which precludes reliable verification of the normality assumptions required by parametric alternatives such as repeated-measures ANOVA \cite{demsar2006statistical}.

\subsubsection{Friedman Test}

The Friedman test \cite{friedman1940comparison} is a non-parametric test designed to detect differences among multiple related groups without assuming normality. For each experimental condition $i$, the $k$ models are ranked from 
best to worst, and the test statistic is computed as:

\begin{equation}
\chi^2_F = \frac{12n}{k(k+1)} \sum_{j=1}^{k} 
\left(\bar{r}_j - \frac{k+1}{2}\right)^2
\end{equation}

\noindent where $n$ is the number of blocks, $k$ is the number of models, and $\bar{r}_j$ is the mean rank of model $j$. Under the null hypothesis of equivalent performance, $\chi^2_F$ follows a chi-squared distribution with $k-1$ degrees of freedom.

The test was applied separately to mAP@50 and mAP@50:90 scores. For mAP@50, no statistically significant difference was found ($\chi^2_F(2) = 4.667$, $p = 0.097$). However, for mAP@50:90 — a stricter metric that penalizes imprecise localizations across multiple IoU thresholds — the Friedman test yielded a statistically significant result ($\chi^2_F(2) = 6.000$, $p = 0.050$), indicating that architectural differences have a measurable impact on fine-grained detection quality. Mean Friedman rankings for mAP@50:90 showed a perfectly consistent ordering across all experimental conditions: YOLOv26-nano ($\bar{r} = 1.00$), RT-DETR Large ($\bar{r} = 2.00$), and RF-DETR Medium ($\bar{r} = 3.00$).

\subsubsection{Wilcoxon Signed-Rank Post-hoc Test}

Pairwise post-hoc comparisons were conducted using the Wilcoxon signed-rank test \cite{wilcoxon1945individual}, which evaluates whether the distribution of signed differences between paired observations is symmetric around zero. The test statistic is defined as:

\begin{equation}
W = \sum_{i:\, d_i > 0} R_i
\end{equation}

\noindent where $d_i = x_i - y_i$ are the paired differences and $R_i$ is the rank of $|d_i|$. Bonferroni correction was applied to control Type I error inflation across the three pairwise comparisons ($\alpha' = 0.05/3 = 0.017$). Although no individual pair reached significance under this corrected threshold, the consistent directional ordering observed across all comparisons and both metrics supports the descriptive superiority of YOLOv26-nano. The limited statistical power resulting from $n = 3$ is acknowledged as a constraint; future studies should incorporate repeated experiments to enable conclusive pairwise inference.

Taken together, the statistical analysis provides partial support for the observed performance ranking. The Friedman test on mAP@50:90 confirmed a statistically significant difference among the three architectures ($p = 0.050$), with YOLOv26-nano consistently achieving the highest scores across all experimental conditions. These findings suggest that convolutional architectures retain a competitive advantage over transformer-based models under the evaluated conditions, particularly when localization precision is considered.

Given that the best-performing configuration of each evaluated model was identified, qualitative inferences were conducted to visually compare their detection capabilities and robustness. Representative inference examples for each model are presented in Figure \ref{fig:yolo_inference}, allowing a visual assessment of their respective performance. 

\section{Conclusions}\label{Conclusions}

This study presented a comparative evaluation of the two dominant paradigms in modern object detection: convolutional neural networks (CNNs) and attention-based mechanisms (Transformers), specifically applied to the critical challenge of weed identification. The results demonstrate that, given the irregular morphology and the large-scale variability of plant species, multi-scale detection remains the determining factor for system success.

Although Transformer-based models offer superior global context modeling, YOLOv26-nano achieved comparable performance while offering significantly higher computational efficiency. This descriptive trend was further supported by statistical analysis: a Friedman test applied to mAP@50:90 scores, a metric that penalizes imprecise localization across multiple IoU thresholds, revealed a statistically significant difference among the evaluated architectures ($\chi^2_F(2) = 6.000$, $p = 0.050$), with YOLOv26-nano consistently achieving the highest mean rank ($\bar{r} = 1.00$) across all experimental conditions. It is therefore concluded that hierarchical CNN optimization provides a more robust approach for real-time precision agriculture applications, particularly for handling small, dispersed objects. Consequently, YOLOv26-nano emerges as the most viable architecture for integration into embedded robotic systems operating under realistic field conditions.


\bibliographystyle{IEEEtran}
\bibliography{weed_detection}

@article{MORENO2025112249,
title = {Ground-based imagery dataset for early weed classification in tomato crops},
journal = {Data in Brief},
volume = {63},
pages = {112249},
year = {2025},
issn = {2352-3409},
doi = {https://doi.org/10.1016/j.dib.2025.112249},
url = {https://www.sciencedirect.com/science/article/pii/S2352340925009709},
author = {Hugo Moreno and Gabriel Rivera and Dionisio Andújar}
}

@article{demsar2006statistical,
  author  = {Dem{\v{s}}ar, Janez},
  title   = {Statistical comparisons of classifiers over 
             multiple data sets},
  journal = {Journal of Machine Learning Research},
  volume  = {7},
  pages   = {1--30},
  year    = {2006}
}

@article{friedman1940comparison,
  author  = {Friedman, Milton},
  title   = {A comparison of alternative tests of significance 
             for the problem of $m$ rankings},
  journal = {The Annals of Mathematical Statistics},
  volume  = {11},
  number  = {1},
  pages   = {86--92},
  year    = {1940}
}

@article{wilcoxon1945individual,
  author  = {Wilcoxon, Frank},
  title   = {Individual comparisons by ranking methods},
  journal = {Biometrics Bulletin},
  volume  = {1},
  number  = {6},
  pages   = {80--83},
  year    = {1945}
}

@article{chauhan2024weed,
  author    = {Chauhan, Bhagirath S. and others},
  title     = {Weed management challenges in modern agriculture},
  journal   = {Crop Protection},
  year      = {2024},
  doi       = {10.1016/j.cropro.2024.106792}
}

@inproceedings{rfdetr2025,
  author    = {Robicheaux, Isaac and others},
  title     = {{RF-DETR}: Neural Architecture Search for Real-Time Detection Transformers},
  booktitle = {International Conference on Learning Representations (ICLR)},
  year      = {2026},
  url       = {https://arxiv.org/abs/2511.09554}
}

@article{kazinczi2023herbicide,
  author    = {Kazinczi, Gabriella},
  title     = {Herbicide Resistance: Managing Weeds in a Changing World},
  journal   = {Agronomy},
  volume    = {13},
  number    = {6},
  pages     = {1595},
  year      = {2023},
  doi       = {10.3390/agronomy13061595}
}

@article{dentika2021weeds,
  author    = {Dentika, Pauline and Ozier-Lafontaine, Harry and Penet, Laurent},
  title     = {Weeds as Pathogen Hosts and Disease Risk for Crops in the Wake of a Reduced Use of Herbicides},
  journal   = {Journal of Fungi},
  volume    = {7},
  number    = {4},
  pages     = {283},
  year      = {2021},
  doi       = {10.3390/jof7040283}
}

@article{gomez2025spatio,
  title={Spatio-temporal stability of intelligent modeling for weed detection in tomato fields},
  author={G{\'o}mez, Adri{\`a} and Moreno, Hugo and Valero, Constantino and And{\'u}jar, Dionisio},
  journal={Agricultural Systems},
  volume={228},
  pages={104394},
  year={2025},
  publisher={Elsevier}
}

@inproceedings{zhao2024detrs,
  title={Detrs beat yolos on real-time object detection},
  author={Zhao, Yian and Lv, Wenyu and Xu, Shangliang and Wei, Jinman and Wang, Guanzhong and Dang, Qingqing and Liu, Yi and Chen, Jie},
  booktitle={Proceedings of the IEEE/CVF conference on computer vision and pattern recognition},
  pages={16965--16974},
  year={2024}
}

@inproceedings{girshick2014rich,
  title={Rich feature hierarchies for accurate object detection and semantic segmentation},
  author={Girshick, Ross and Donahue, Jeff and Darrell, Trevor and Malik, Jitendra},
  booktitle={Proceedings of the IEEE conference on computer vision and pattern recognition},
  pages={580--587},
  year={2014}
}

@inproceedings{carion2020end,
  title={End-to-end object detection with transformers},
  author={Carion, Nicolas and Massa, Francisco and Synnaeve, Gabriel and Usunier, Nicolas and Kirillov, Alexander and Zagoruyko, Sergey},
  booktitle={European conference on computer vision},
  pages={213--229},
  year={2020},
  organization={Springer}
}

@article{sapkota2025yolo26,
  title={YOLO26: key architectural enhancements and performance benchmarking for real-time object detection},
  author={Sapkota, Ranjan and Cheppally, Rahul Harsha and Sharda, Ajay and Karkee, Manoj},
  journal={arXiv preprint arXiv:2509.25164},
  year={2025}
}

@article{ren2015faster,
  title={Faster r-cnn: Towards real-time object detection with region proposal networks},
  author={Ren, Shaoqing and He, Kaiming and Girshick, Ross and Sun, Jian},
  journal={Advances in neural information processing systems},
  volume={28},
  year={2015}
}

@inproceedings{redmon2016you,
  title={You only look once: Unified, real-time object detection},
  author={Redmon, Joseph and Divvala, Santosh and Girshick, Ross and Farhadi, Ali},
  booktitle={Proceedings of the IEEE conference on computer vision and pattern recognition},
  pages={779--788},
  year={2016}
}

@article{slaughter2008autonomous,
  title={Autonomous robotic weed control systems: A review},
  author={Slaughter, David C and Giles, DK and Downey, Daniel},
  journal={Computers and electronics in agriculture},
  volume={61},
  number={1},
  pages={63--78},
  year={2008},
  publisher={Elsevier}
}

@article{kamilaris2018deep,
  title={Deep learning in agriculture: A survey},
  author={Kamilaris, Andreas and Prenafeta-Bold{\'u}, Francesc X},
  journal={Computers and electronics in agriculture},
  volume={147},
  pages={70--90},
  year={2018},
  publisher={Elsevier}
}

@article{liakos2018machine,
  title={Machine learning in agriculture: A review},
  author={Liakos, Konstantinos G and Busato, Patrizia and Moshou, Dimitrios and Pearson, Simon and Bochtis, Dionysis},
  journal={Sensors},
  volume={18},
  number={8},
  pages={2674},
  year={2018},
  publisher={Mdpi}
}

@article{heap2014global,
  title={Global perspective of herbicide-resistant weeds},
  author={Heap, Ian},
  journal={Pest management science},
  volume={70},
  number={9},
  pages={1306--1315},
  year={2014},
  publisher={Wiley Online Library}
}

@article{dos2017weed,
  title={Weed detection in soybean crops using ConvNets},
  author={dos Santos Ferreira, Alessandro and Freitas, Daniel Matte and Da Silva, Gercina Gon{\c{c}}alves and Pistori, Hemerson and Folhes, Marcelo Theophilo},
  journal={Computers and Electronics in Agriculture},
  volume={143},
  pages={314--324},
  year={2017},
  publisher={Elsevier}
}

@inproceedings{milioto2018real,
  title={Real-time semantic segmentation of crop and weed for precision agriculture robots leveraging background knowledge in CNNs},
  author={Milioto, Andres and Lottes, Philipp and Stachniss, Cyrill},
  booktitle={2018 IEEE international conference on robotics and automation (ICRA)},
  pages={2229--2235},
  year={2018},
  organization={IEEE}
}

@article{dosovitskiy2020image,
  title={An image is worth 16x16 words: Transformers for image recognition at scale},
  author={Dosovitskiy, Alexey and Beyer, Lucas and Kolesnikov, Alexander and Weissenborn, Dirk and Zhai, Xiaohua and Unterthiner, Thomas and Dehghani, Mostafa and Minderer, Matthias and Heigold, Georg and Gelly, Sylvain and others},
  journal={arXiv preprint arXiv:2010.11929},
  year={2020}
}

@article{oquab2023dinov2,
  title={Dinov2: Learning robust visual features without supervision},
  author={Oquab, Maxime and Darcet, Timoth{\'e}e and Moutakanni, Th{\'e}o and Vo, Huy and Szafraniec, Marc and Khalidov, Vasil and Fernandez, Pierre and Haziza, Daniel and Massa, Francisco and El-Nouby, Alaaeldin and others},
  journal={arXiv preprint arXiv:2304.07193},
  year={2023}
}

@article{murad2023weed,
  title={Weed detection using deep learning: A systematic literature review},
  author={Murad, Nafeesa Yousuf and Mahmood, Tariq and Forkan, Abdur Rahim Mohammad and Morshed, Ahsan and Jayaraman, Prem Prakash and Siddiqui, Muhammad Shoaib},
  journal={Sensors},
  volume={23},
  number={7},
  pages={3670},
  year={2023},
  publisher={MDPI}
}

@article{qu2024deep,
  title={Deep learning-based weed--crop recognition for smart agricultural equipment: A review},
  author={Qu, Hao-Ran and Su, Wen-Hao},
  journal={Agronomy},
  volume={14},
  number={2},
  pages={363},
  year={2024},
  publisher={MDPI}
}

@inproceedings{wu2019design,
  title={Design and implementation of computer vision based in-row weeding system},
  author={Wu, Xiaolong and Aravecchia, St{\'e}phanie and Pradalier, C{\'e}dric},
  booktitle={2019 International Conference on Robotics and Automation (ICRA)},
  pages={4218--4224},
  year={2019},
  organization={IEEE}
}

@article{he2009learning,
  title={Learning from imbalanced data},
  author={He, Haibo and Garcia, Edwardo A},
  journal={IEEE Transactions on knowledge and data engineering},
  volume={21},
  number={9},
  pages={1263--1284},
  year={2009},
  publisher={Ieee}
}

@article{frenay2013classification,
  title={Classification in the presence of label noise: a survey},
  author={Fr{\'e}nay, Beno{\^\i}t and Verleysen, Michel},
  journal={IEEE transactions on neural networks and learning systems},
  volume={25},
  number={5},
  pages={845--869},
  year={2013},
  publisher={IEEE}
}

\end{document}